\documentclass[10pt,twocolumn,letterpaper]{article}
\usepackage{hyperref}
\pdfoutput=1

\usepackage[USenglish]{babel}
\usepackage{times}
\usepackage{graphicx}
\usepackage{amsmath}
\usepackage{amssymb}
\usepackage{iccv}

\usepackage{graphicx,subfigure,color,epstopdf}
\usepackage{todonotes}
\usepackage{etoolbox}
\usepackage{booktabs}

\iccvfinalcopy
\begin{document}

\title{ 
Training a Convolutional Neural Network for \\ Appearance-Invariant Place Recognition
}

\author{
\begin{tabular}[c]{c @{\extracolsep{1em}} c @{\extracolsep{1em}} c}
        Ruben Gomez-Ojeda$^{1}$ &
        Manuel Lopez-Antequera$^{1,2}$ \\
        Nicolai Petkov$^{2}$ &
        Javier Gonzalez-Jimenez$^{1}$ 
        \\
\end{tabular}
\cr
\cr
\small
\begin{tabular}[c]{c@{\extracolsep{4em}}c} 
       $^1$ University of M\'alaga & $^2$ University of Groningen
\end{tabular}
}


\maketitle

\begin{abstract}
Place recognition is one of the most challenging problems in computer vision, and has become a key part in mobile robotics and autonomous driving applications for performing loop closure in visual SLAM systems. Moreover, the difficulty of recognizing a revisited location increases with appearance changes caused, for instance, by weather or illumination variations, which hinders the long-term application of such algorithms in real environments. In this paper we present a convolutional neural network (CNN), trained for the first time with the purpose of recognizing revisited locations under severe appearance changes, which maps images to a low dimensional space where Euclidean distances represent place dissimilarity. In order for the network to learn the desired invariances, we train it with triplets of images selected from datasets which present a challenging variability in visual appearance. The triplets are selected in such way that two samples are from the same location and the third one is taken from a different place. We validate our system through extensive experimentation, where we demonstrate better performance than state-of-art algorithms in a number of popular datasets.
\end{abstract}

\newcommand {\indfig}{0mm}				%
\newcommand {\indtab}{2mm}				%
\newcommand {\indleg}{5mm}				%

\newcommand{\wintro}     	{0.475\textwidth}	%
\newcommand{\warch}     	{0.70\textwidth}	%
\newcommand{\wkitti}     	{0.40\textwidth}	%
\newcommand{\walderley}  	{0.35\textwidth}	%
\newcommand{\wmalaga}    	{0.35\textwidth}	%

\newcommand{\wkitticonfmat}  	{0.29\textwidth}	%
\newcommand{\wkitticonfbin}  	{0.29\textwidth}	%

\newcommand{\wkitticurv}  	{0.35\textwidth}	%
\newcommand{\wmalagacurv}    	{0.35\textwidth}	%
\newcommand{\wnordcurv}  	{0.35\textwidth}	%
\newcommand{\waldcurv}  	{0.35\textwidth}	%

\newcommand{\wlegend}  		{0.35\textwidth}	%

\renewcommand{\arraystretch}{1.2} %

\newcommand{\img}{\textbf{x}}				%
\newcommand{\imgi}{\textbf{x}_i}			%
\newcommand{\imgj}{\textbf{x}_j}			%
\newcommand{\imgk}{\textbf{x}_k}			%
\newcommand{\imgw}{M}					%
\newcommand{\imgh}{N}					%
\newcommand{\imgd}{C}					%
\newcommand{\descl}{D}					%
\newcommand{\cnnmap}{\textbf{h}}			%
\newcommand{\desci}{\cnnmap ( \imgi )}			%
\newcommand{\descj}{\cnnmap ( \imgj )}			%
\newcommand{\desck}{\cnnmap ( \imgk )}			%
\newcommand{\gcnn}{CaffeNet }				%

\newcommand{\objfun}{\mathcal{L}}			%
\newcommand{\cnnpar}{\bs\omega}				%
\newcommand{\costfun}{\mathcal{C}}			%
\newcommand{\margin}{\beta}				%
\newcommand{\marginval}{1}				%
\newcommand{\confmat}{\mathcal{M}}			%

\newcommand{\triplet}{t}				%
\newcommand{\trainset}{\bs{\tau}_{}}			%
\newcommand{\testset}{\bs{\tau}_{}}			%
\newcommand{\fig}[1]{Figure \ref{#1}}					%
\newcommand{\figs}[2]{Figures \ref{#1} and \ref{#2}}			%
\newcommand{\tab}[1]{Table \ref{#1}}					%
\newcommand{\secref}[1]{Section {\ref{#1}}}				%

\newcommand{\bs}[1]{\boldsymbol{#1}}					%
\newcommand{\ssl}[1]{\tensor[^{#1}]}					%
\newcommand{\MatrixS}[1]{\bs{#1}}					%
\newcommand{\MatrixL}[1]{\textbf{#1}}					%
\newcommand{\brackets}[1]{\begin{bmatrix}#1\end{bmatrix}}		%

\newcommand{\argmin}[1]{\underset{#1}{\operatorname{argmin}}}		%
\newcommand{\argmax}[1]{\underset{#1}{\operatorname{argmax}}}		%
\newcommand{\der}[2]{\frac{\partial #1}{\partial #2}}			%
\newcommand{\derin}[3]{\left.\der{#1}{#2}\right|_{#3}} 			%
\newcommand{\prob}[2]{p\left( #1 | #2 \right)}				%
\newcommand{\norm}[1]{\left\lVert#1\right\rVert} 			%
\newcommand{\eucnorm}[1]{\left\lVert#1\right\rVert_{2}}			%
\newcommand{\fnorm}[1]{\left\lVert#1\right\rVert_{\mathfrak{F}}}	%
\newcommand{\skewmat}[1]{ \left[#1\right]_\times }			%

\newcommand{\IdMat}{\MatrixL{I}}					%
\newcommand{\canonicalvec}[1]{\textbf{e}_{#1}}				%
\newcommand{\TRANSPOSE}{^\top}						%
\newcommand{\symcov}{\MatrixS{\Sigma}}					%
\newcommand{\symRe}{\mathbb{R}}						%
\newcommand{\symSSpace}{S}						%
\newcommand{\symPSpace}{\mathbb{P}}					%
\newcommand{\symRotSpace}{SO(3)}					%
\newcommand{\symRotLie}{\mathfrak{so}(3)}				%
\newcommand{\symEucSpace}{SE(3)}					%
\newcommand{\symEucLie}{\mathfrak{se}(3)}				%
\newcommand{\symrot}{\textbf{R}}					%
\newcommand{\symtrans}{\textbf{t}}					%

\newcommand{\idL}{L}				%
\newcommand{\idR}{R}				%
\newcommand{\idF}{k}				%
\newcommand{\idFn}{k+1}				%

\newcommand{\lIm}{\textbf{l}}			%
\newcommand{\normL}{\eta_l}			%
\newcommand{\lImFirst}{\lIm_{\idL,\idF}}	%
\newcommand{\lImSecond}{\lIm_{\idR,\idF}}	%
\newcommand{\lImThird}{\lIm_{\idL,\idFn}}	%
\newcommand{\lImFourth}{\lIm_{\idR,\idFn}}	%

\newcommand{\match}{\textbf{m}}			%

\newcommand{\cam}{C}				%
\newcommand{\calib}{\MatrixL{K}}		%
\newcommand{\LO}{LO}				%
\newcommand{\spoint}{\textbf{p}}		%
\newcommand{\epoint}{\textbf{q}}		%
\newcommand{\Spoint}{\textbf{P}}		%
\newcommand{\Epoint}{\textbf{Q}}		%
\newcommand{\spointx}{p_x}		%
\newcommand{\spointy}{p_y}		%

\newcommand{\separam}{\bs{\xi}}
\newcommand{\separaminc}{\bs{\varepsilon}}
\newcommand{\separamopt}{\separam^*}
\newcommand{\reltrans}{\MatrixL{T}(\separam)}
\newcommand{\reltransopt}{\MatrixL{T}(\separamopt)}

\newcommand{\symover}{\gamma}
\newcommand{\errfun}{\MatrixL{E}}
\newcommand{\weifun}{\MatrixL{W}}
\newcommand{\jacfun}{\MatrixL{J}}

\newcommand{\cross}{\times}	%
\newcommand{\lx}{(p_y-q_y)}
\newcommand{\ly}{(q_x-p_x)}
\newcommand{\lbeta}{(p_x^2+p_y^2+q_x^2+q_y^2Id)-2(p_xq_x+p_yq_y}

\section{Introduction}
\begin{figure}[!ht]
  \centering
    \includegraphics[width=\wintro]{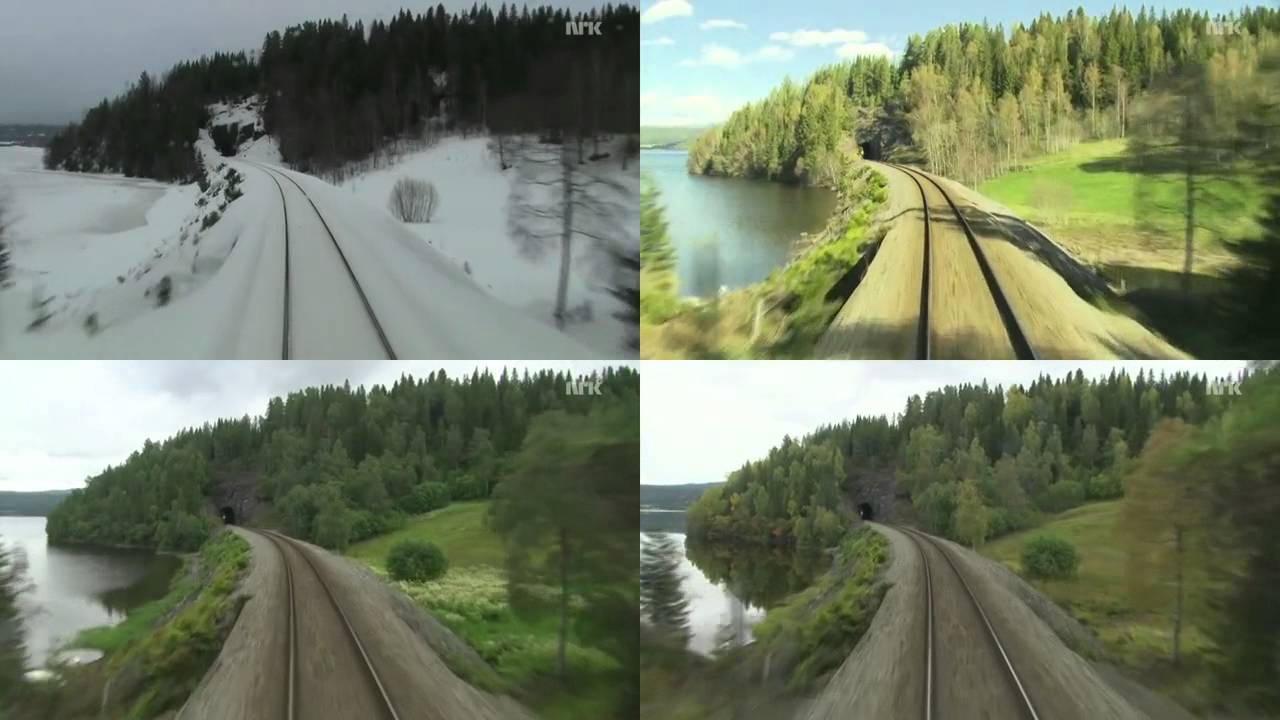}
    \caption{
    Frames extracted from the Nordland dataset \cite{Sunderhauf2013} that belong to the same place in winter, spring, summer and fall. The proposed method is capable of recognizing the same location under challenging appearance changes.
    }
    \vspace{\indfig}
    \label{fig_nordland}
\end{figure}
The process of identifying images that belong to the same location, usually known as place recognition, is still an open problem in computer vision.
Place recognition is a key part in mobile robotics and autonomous driving applications, such as vision-based simultaneous localization and mapping (SLAM) systems, where revisiting a location introduces important information which can be employed in the tasks of localization \cite{rivera2015appearance} and loop closure \cite{Mur-Artal2014}.
It also can be applied in augmented reality applications, where the user obtains information about important places, monuments or texts from a single image taken with a smartphone camera.
The difficulties induced by changes in the scenario, viewpoint, illumination or weather conditions makes place recognition a much more difficult task than one may intuitively think (see \fig{fig_nordland}).
Traditionally, place recognition has focused on scenarios without major appearance changes. In that context, most methods employ bags of visual words inspired by \cite{Sivic2003} and \cite{Nister2006}. Bag-of-words (BoW) approaches have proven to work quickly and effectively in static scenes, but they have several drawbacks.
They usually rely on traditional keypoint descriptors, such as SIFT \cite{lowe2004sift}, SURF \cite{Bay2006}, or BRIEF \cite{Calonder2010}, which describe the local appearance of individual patches, limiting their descriptive power with respect to whole image methods, as observed by \cite{Milford2012}. Their performance in challenging environments strongly depends on the invariance of those descriptors to perceptual changes. %
Convolutional neural networks (CNNs) are gaining importance in most classification tasks \cite{Krizhevsky2012}. When used as generic feature generators, they often outperform the state-of-art algorithms even for tasks different to classification \cite{Sharif2014}. However, their use in place recognition is limited to the exploitation of generic features extracted from the internal layers of pre-trained CNNs \cite{Chen2014}\cite{Sunderhauf2015}.

In this paper, we propose a novel approach to place recognition capable of detecting revisited places under extreme changes in weather, illumination, or external conditions. In contrast to previous algorithms which rely on visual descriptors, our algorithm works with the complete image, reducing unnecessary errors induced by posterior feature matching processes by providing a better estimate of place similarity.
For that purpose, we have trained a CNN for the task of recognizing revisited places. To the best of our knowledge, this is the first CNN specifically trained to perform place recognition as opposed to using generic features extracted from networks trained for other tasks. 
We demonstrate that place recognition can be better resolved by discriminatively training a network for such a problem, since visual cues that are relevant for object classification may not be optimal for place recognition.
Moreover, we claim that place recognition can be performed with a smaller network than those employed for object recognition.
We contribute to the state of the art with a CNN:
\begin{itemize}
\item[$\circ$] Capable of recognizing revisited places under challenging appearance changes of the scene, including seasonal, time of day and outdoor/indoor changes.
\item[$\circ$] Suitable for any long term, real time place recognition tasks which are often necessary in mobile robotics and autonomous navigation.
\end{itemize}
We demonstrate these claims with extensive experimentation in several challenging datasets, where we compare our proposal with two state-of-art algorithms: DBoW2 \cite{mur2015orb}, and a generic network as in \cite{Pepperell2014}. Experiments show the better performance of our method, which recognizes previously visited locations under severe appearance changes with a higher rate of success than the state-of-art algorithms, with an inferior computational burden than previous CNN-based methods on datasets where appearance changes are severe.

\section{Related Work}
As mentioned above, visual place recognition has been object of research under the field of SLAM, often as a key part of the localization and loop closing modules. %
One of the first SLAM techniques which introduced BoW in this context was FAB-MAP \cite{Cummins2008}, where a probabilistic approach to place recognition based on the local appearance of each location was proposed. They also deal with perceptual aliasing in the environment by introducing a generative model which implements some logic reasoning to discard false positives caused by this phenomena. However, the use of SURF features and the employment of the generative model increases the computational burden.
This was tackled in \cite{galvez2012bags} with DBoW2, where for the first time they introduced the use bags of binary words obtained from BRIEF descriptors, reducing in more than an order of magnitude the time employed in the feature extraction process. The use of BRIEF, which is not rotation or scale invariant, limits the recognition task to scenes taken from the same viewpoint in planar trajectories. An improved version of this algorithm has been recently published in \cite{mur2015orb}, where the authors build a urban dictionary based on ORB \cite{rublee2011orb} which yields a better performance in popular datasets.

A common problem to previous techniques is their poor behavior in place recognition under different illumination conditions and poorly textured environments, and also their limited invariance to scale and viewpoint.
In \cite{Lee2014}, the authors deal with that by building a vocabulary tree that employs straight lines in combination with the MSLD descriptor \cite{Wang2009}, which increases the robustness against changes in weather conditions. However, the evaluation sequences do not include strong perceptual changes, thus the system may not be suitable to long-term operations in changing environments. %
This problem was tackled by Neubert et al. in \cite{Neubert2013a}, where they propose a place recognition algorithm capable of working across seasons. They argue that seasonal changes in the scene are predictable, and propose a superpixel-based algorithm (SP-APC) which is able to predict those changes and then recognize the scene, with a prediction process based on a dictionary that learns from training data how the appearance of the scene changes over the year. On the other hand, the algorithm is only tested with the Nordland dataset \cite{Sunderhauf2013}, which shows extreme seasonal changes, and hence it will not predict gradual changes in the environment.
A different strategy works on local sequences instead of estimating the best single location, with the proposal of Milford and Wyeth as one of the most relevant contributions \cite{Milford2012}. They propose SeqSLAM, a post-processing technique that recognizes sequences of locations previously visited, under challenging perceptual changes. Their approach estimates the best match by taking into account not only the single location, but also imposing coherence with the surrounding sequence. Under this assumption, they obtain a good performance by only applying a local contrast enhancement to the input images (downsampled from the original datasets), and then comparing the normalized images by processing the sum of absolute differences (SAD) between them. However, this procedure has several drawbacks. It only works with local and consistent sequences, which makes it impractical for applications that work with isolated images. It also may fail with big changes of viewpoint and rotation, and also suffers image aliasing since its viewpoint invariance is only due to extreme downscaling of the input images.
Recently, another group of techniques has irrupted with promising results, motivated by the outstanding performance achieved by CNNs as generic feature generators in several classification tasks \cite{Sharif2014}. In this context, 
a recent work is \cite{Sunderhauf2015}, where the authors employ a pre-trained network named OverFeat \cite{Sermanet2013}, which was the winner of the localization task of the ImageNet Large Scale Visual Recognition Challenge 2013 \cite{russakovsky2014imagenet}. They study the use of the intermediate representations learned by the CNN as image features valuable for place recognition even under challenging appearance changes, with promising results.
\begin{figure*}[!htb]
	\centering	
		\includegraphics[width=\warch]{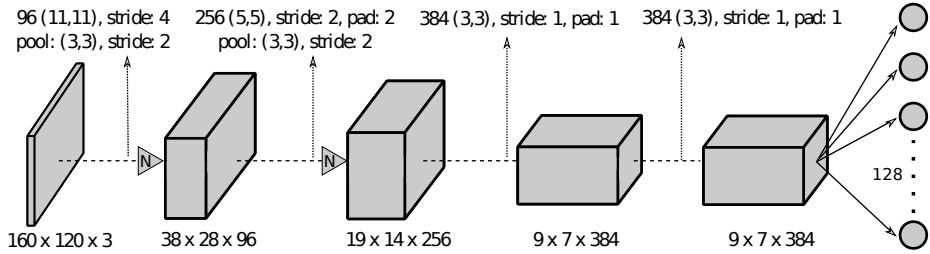}
  	\caption{ 
  	Architecture of the proposed network. The convolution and pooling stages
  	are indicated at the top of the figure, and the sizes of the resulting data
  	are shown on the bottom part. $\mathrm{N}$ is a local contrast normalization operation  
  	acting across channels as applied in \cite{Krizhevsky2012}.
	} 
	\vspace{\indfig}
  	\label{fig_netarch}
\end{figure*}
\section{Methodology}
To solve the task of detecting if an image
belongs to a previously visited place, 
we propose to train a Convolutional Neural Network
to embed images in a low dimensional space where Euclidean distance
represents location dissimilarity. 
Our solution is inspired by works in content-based image retrieval,
however, our network is trained to produce a feature vector invariant to drastic appearance changes in
the scene such as seasonal changes.
To  achieve this, we train the network using labeled
datasets which present the same locations under different
illumination, point of view or weather conditions. We apply a training technique similar to \cite{Wohlhart2015}, 
where the network presented with triplets of images, formed by
a query image $\imgi$, an image from the same location, $\imgj$,
and an image of a different location, $\imgk$. In the following we describe the architecture and 
the training process of the proposed network.

\subsection{Architecture of the CNN}
In view of the difficulties in training a convolutional neural 
network from scratch using a relatively small specialized 
dataset, we take the approach of modifying a pre-trained network.
In particular, we resort to the reference CaffeNet network 
\cite{jia2014caffe}, which mirrors the architecture of Krizhevsky 
et al. \cite{Krizhevsky2012}, from which we only keep the first 
four convolutional layers, replacing 
the rest with a single fully connected layer which is our 
descriptor output (see \fig{fig_netarch}). Since we discard all the fully connected 
layers, we are not constrained to the original input size 
of $227 \times 227$ pixels and instead work with a smaller input 
of $160 \times 120$. 

\subsection{Description of the Cost Function}
\label{cfdescript}
In a nutshell, the network maps a $\imgw \times \imgh \times \imgd$ input image to a descriptor vector of length $\descl$, which corresponds to the activations of the output layer of the CNN, i.e.:
\begin{align}
\cnnmap: & \; \; \mathbb{R}^{\imgw \times \imgh \times \imgd} \; \; \longmapsto \; \; \mathbb{R}^{\descl}  			\nonumber \\
		 & \; \; \; \; \; \; \; \img \; \; \; \; \; \; \; \;  \; \; \longmapsto \; \; \cnnmap ( \img )
\end{align}
being $\cnnmap ( \img )$ the descriptor of the image $\img$, whose Euclidean distances to other descriptors must be representative of location dissimilarity. 
In order to achieve this behavior, the network parameters $\cnnpar^*$ are obtained by minimizing the following objective function
\begin{equation}
\cnnpar^* =  \argmin{\cnnpar} \Big\{  \; \objfun + \lambda \eucnorm{\cnnpar} ^2 \Big\}
\end{equation}
where the second term represents a regularization over the parameters of the network $\cnnpar$, and the first term $\objfun$ is the sum of the cost functions over all the triplets, that can be expressed as
\begin{equation}
\objfun = \sum_{(\imgi,\imgj,\imgk)\in\trainset} \costfun(\imgi,\imgj,\imgk)
\end{equation} 
with $\costfun$ being the cost function for each triplet of images. The cost function employed is similar to that in \cite{Wohlhart2015}, and can be expressed as:
\begin{equation}
\costfun(\imgi,\imgj,\imgk) = max \Big\{ 0,1-\frac{\eucnorm{\desci-\desck}}{\margin+\eucnorm{\desci-\descj}}\Big\}
\end{equation}
This cost function is satisfiable when the distance of the 
dissimilar pair is larger than the distance of the similar pair 
by at least a margin $\margin$, producing zero cost.
This means that dissimilar descriptors will not continue to be 
separated indefinitely in the descriptor space during training.
On the contrary, triplets not satisfying this condition will produce costs that the training process will aim to reduce by updating the weights of the CNN accordingly.

\subsection{Training the CNN}
\begin{figure}[t!]
	\centering
	\subfigure[Query image]{	
		\includegraphics[width=\wkitti]{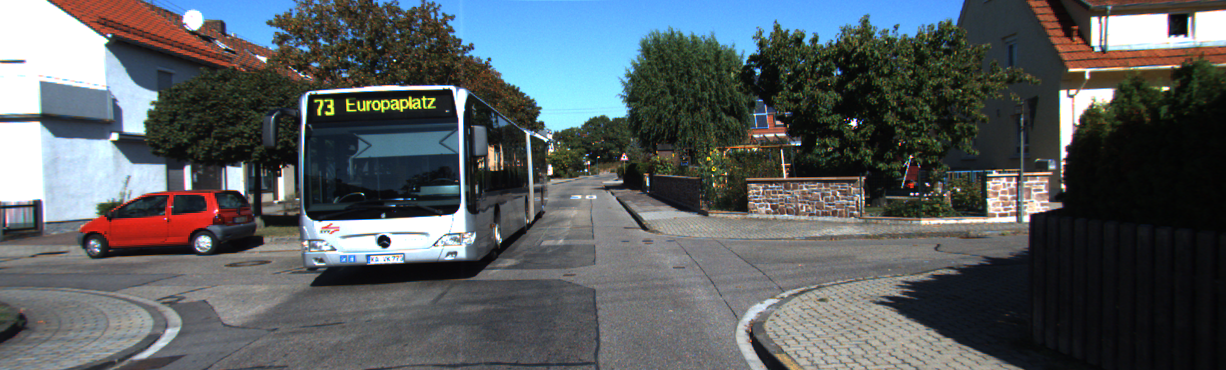}
		\vspace{\indfig}
		\label{fig_kitti_a}
		}
	~
	\subfigure[Similar image]{	
		\includegraphics[width=\wkitti]{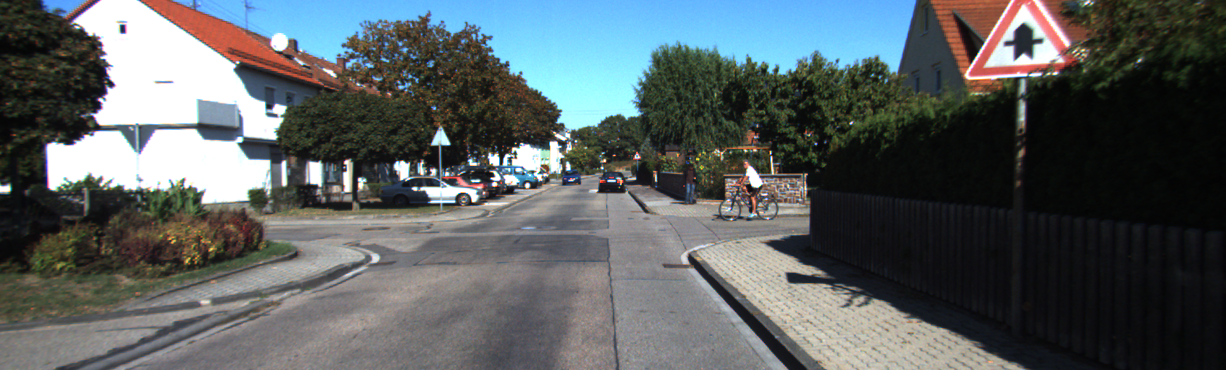}
		\vspace{\indfig}
		\label{fig_kitti_b}
		}	
	~
	\subfigure[Different image]{	
		\includegraphics[width=\wkitti]{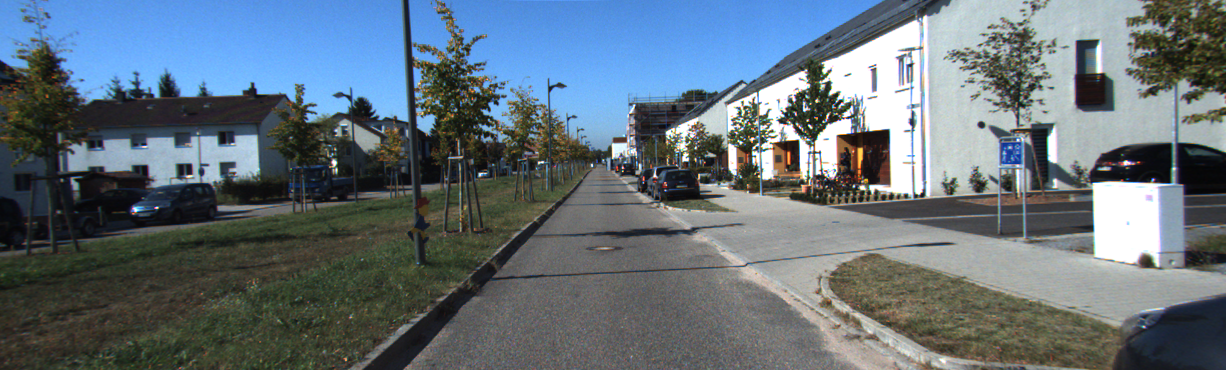}
		\vspace{\indfig}
		\label{fig_kitti_c}
		}	
  	\caption{ 
	Training triplet extracted from the KITTI dataset \cite{geiger2012we}, where large viewpoint invariances are present.	
	} 
	\vspace{\indfig}
  	\label{fig_kitti}
\end{figure}

To achieve the desired invariances in the representation produced by
the network, triplets must be chosen as to
provide relevant visual cues (see \fig{fig_kitti} for an example). 
We train the network using a mixture of triplets from several datasets,
which are detailed in the following sections, to improve invariance
to lighting, weather and point of view changes.
The network is trained using the Caffe library 
\cite{jia2014caffe}, modified to include the previously 
described cost function.  

As previously explained, the weights of the four convolutional layers are fine-tuned from the CaffeNet reference network, an implementation of \cite{Krizhevsky2012}, whereas the final
fully connected layer is new. We scale the learning rate of
the pre-trained layers by a factor of $1/1000$ and fix the
global learning rate at $0.001$. The margin $\margin$ is set to $\marginval$ and the regularization constant $\lambda$ to
$0.0005$. We train for a $40.000$ iterations, for a total
of $1.2$ million triplets, using portions fo the KITTI, Nordland, and Alderley datasets, which are described in the following sections.

\subsubsection{KITTI Dataset} 
The odometry benchmark from the KITTI dataset \cite{geiger2012we} is comprised of 11 training sequences with accurate ground truth of the trajectory, and 10 test sequences without ground truth for evaluation. Both the training and the test sequences are stereo frames extracted from urban environments in daylight conditions. 
We select triplets in order to increase the robustness of the network to changes in viewpoint by choosing the similar pair in a wide variety of relative poses. We also check that the different pairs do not belong to the same place by employing the ground truth location (since loop closures exist in the sequences) 
\fig{fig_kitti} depicts a triplet extracted from the KITTI dataset.

\subsubsection{Alderley Dataset}
\begin{figure}[b!]
	\centering
	\subfigure[Daylight sequence]{	
		\includegraphics[width=\walderley]{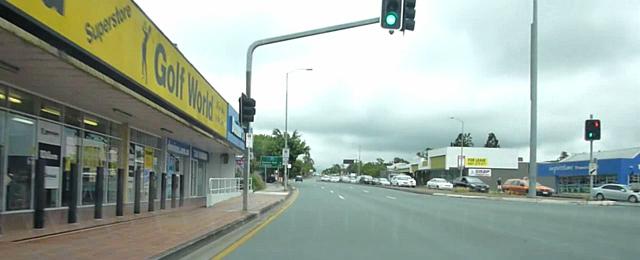}
		\vspace{\indfig}
		\label{fig_alderley_a}
		}
	~
	\subfigure[Stormy-night sequence]{	
		\includegraphics[width=\walderley]{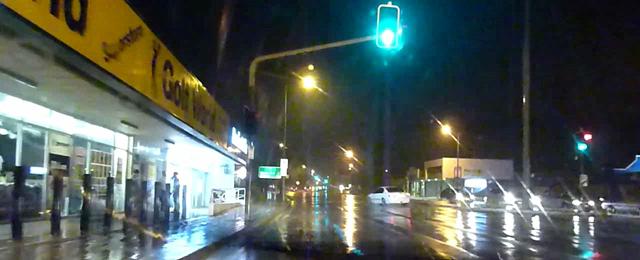}
		\vspace{\indfig}
		\label{fig_alderley_b}
		}	
  	\caption{ 
	Frames extracted from the Alderley dataset \cite{Milford2012}, where drastic illumination changes are present.	
	} 
	\vspace{\indfig}
  	\label{fig_alderley}
\end{figure}

We have also trained the network with the Alderley dataset \cite{Milford2012}, which contains severe changes in illumination and weather conditions. This dataset is formed by two sequences of 8 km along the suburb of Alderley in Brisbane (Australia). The first one was recorded during a clear morning, while the second one was collected in a stormy night with low visibility (see \fig{fig_alderley}). In order to achieve robustness to the aforementioned changes, during training we provide the network with challenging triplets that combine images from both sequences (we have used the first 10k frames from the day sequence and their matches from the night sequence for the training, while reserving the rest for experimentation).

\subsubsection{Nordland Dataset} 
The Nordland dataset \cite{Sunderhauf2013}, extracted from the TV documentary ``Nordlandsbanen - Minutt for Minutt" produced by the Norwegian Broadcasting Corporation NRK consists of a 728 km long train journey connecting the cities of Trondheim and Bod$\o$ in Norway. The sequence was recorded once in each season, and hence it contains challenging appearance changes, as \fig{fig_nordland} shows.
Additionally, it provides different weather conditions due to the large length of the dataset (the sequences are 10 hour long approximately). We generate triplets by providing two images from the same place in different seasons, and an image from another location in any season (we check that frames are actually from different places using the included GPS ground truth). 

\section{Experimental Evaluation}
In order to validate the proposed network, we perform a series of experiments where we compare the behavior of our system with two state-of-art techniques in place recognition: DBoW2 \cite{mur2015orb}, and a feature vector extracted from an internal layer of a neural network trained for object classification as in \cite{Sunderhauf2015}. The actual implementations used are the official distribution of ORB-SLAM \cite{mur2015orb}, and the CaffeNet \cite{jia2014caffe} implementation of \cite{Krizhevsky2012}, which we simply name as \gcnn in this work.
The resolutions of the input images are $160 \times 120$ in our proposal, $227 \times 227$ in \gcnn, and the native resolution of each dataset in DBoW2.
In the following, we first describe the methodology employed for the comparison, then we present a number of experiments with datasets from several environments, under different appearance changes. 
Finally, we also compare the computational cost of the algorithms and their feasibility for place recognition tasks, such as loop closure modules in visual SLAM algorithms.

\subsection{On Comparing Confusion Matrices}
The key element of a place recognition system is the estimation of the similarity between the compared images. For that purpose, we calculate a descriptor $\desci$ for each input image $\imgi$, and then we estimate the similarity with other images by comparing the Euclidean distance from their descriptors. A common measurement widely employed in place recognition collects each distance (or score) in a confusion matrix, where the rows and the columns express the database and the query sequence, respectively, that is $\confmat(i,j) = \eucnorm{\desci-\descj}$. In our case, a normalized confusion matrix $\confmat^*$ can be defined as follows:
\begin{equation}
\confmat^*(i,j) = \frac{\confmat(i,j)}{max\{\confmat(i,j)\}}
\end{equation}
whose terms include the normalized Euclidean distance between the descriptors associated to the $i$ and $j$ images from each sequence. For the methods with which we compare our proposal, the confusion matrices include the proposed normalized scores for each methodology, which are:
\begin{itemize}
\item[$\circ$] DBoW2 \cite{mur2015orb}: the proposed score is 
already normalized, but their approach associates high scores to 
similar images, thus we estimate the complementary matrix before 
the comparison.
\item[$\circ$] \gcnn \cite{Sunderhauf2015}: we extract the 
convolutional layers outputs \textit{conv4}, which present the 
best results for the tested datasets, and compare them using 
Euclidean distance as they propose.
\end{itemize}
Place recognition methods for loop closure generally employ post-processing techniques to find good matches which actually represent the same location in the confusion matrix, usually by looking for sequences of similar frames \cite{Milford2013a} \cite{Pepperell2014}. 
Any method that generates a confusion matrix can benefit from such post-processing techniques, including ours. For this reason, we
perform our experimental comparisons on the ``raw'' confusion matrix.
A problem of using confusion matrices to compare the performance of different methods, is that it is quite difficult to establish a indicator of the quality of a confusion matrix, since there is no ground truth measurement of the place similarity between any two images.
To overcome this issue, we perform a comparison based on synchronized sequences which do not present any loop closures, since in those cases the ground truth pair is placed on the diagonal of the confusion matrix.
In order to generate a quality measurement of a confusion matrix, we start by only keeping its $k$ smallest values. 
Then we plot the ratio of points that fall within the diagonal with respect to $d$, which is defined as the maximum distance to the diagonal to consider a point as an inlier (see \fig{fig_kitticurv}).

\subsection{KITTI Dataset}
\begin{figure*}[!htb]
	\centering
	\subfigure[Our approach]{	
		\includegraphics[width=\wkitticonfmat]{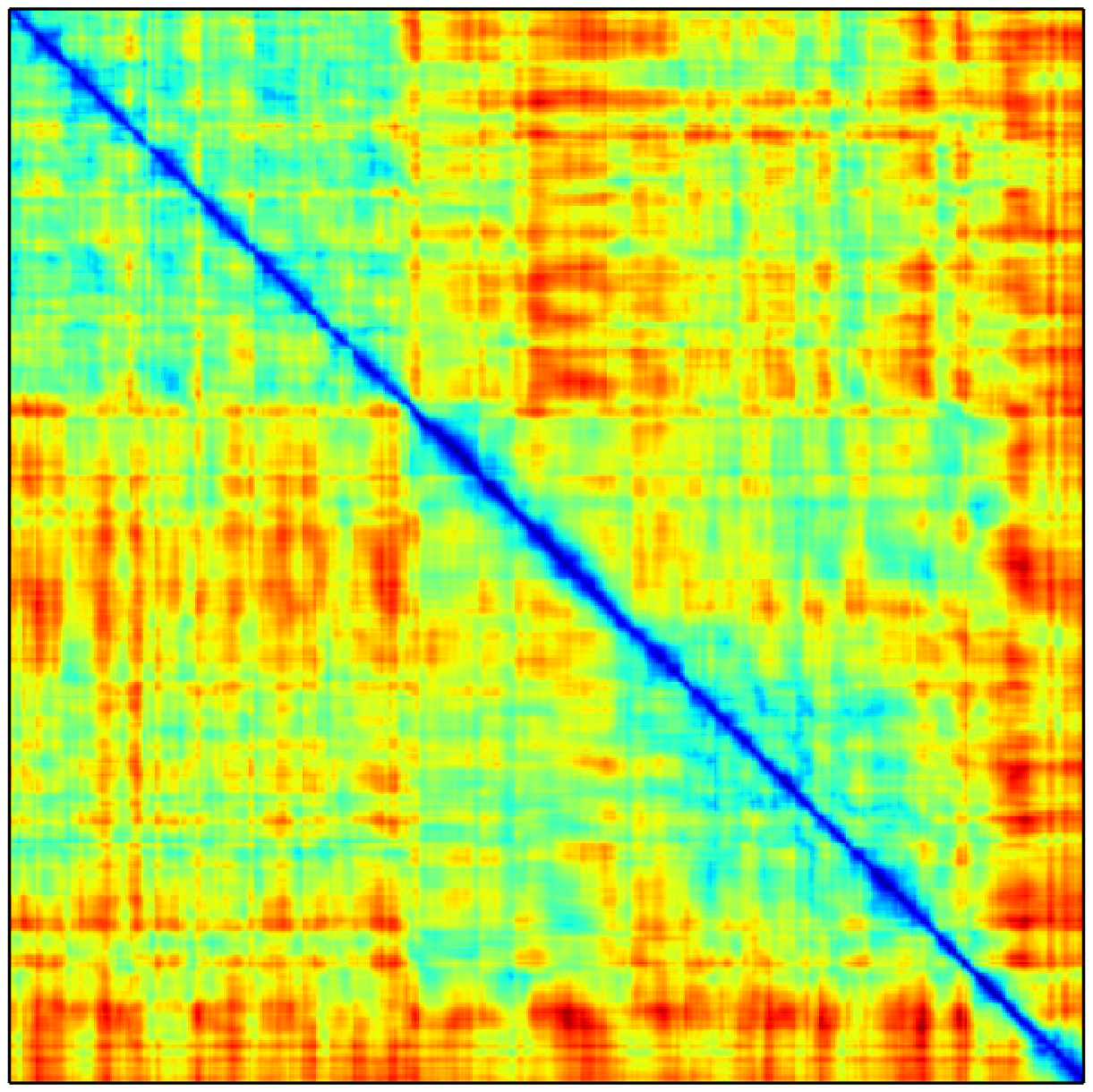}
		\vspace{\indfig}
		\label{fig_kitticonfmat_a}
		}
	~
	\subfigure[DBoW2]{	
		\includegraphics[width=\wkitticonfmat]{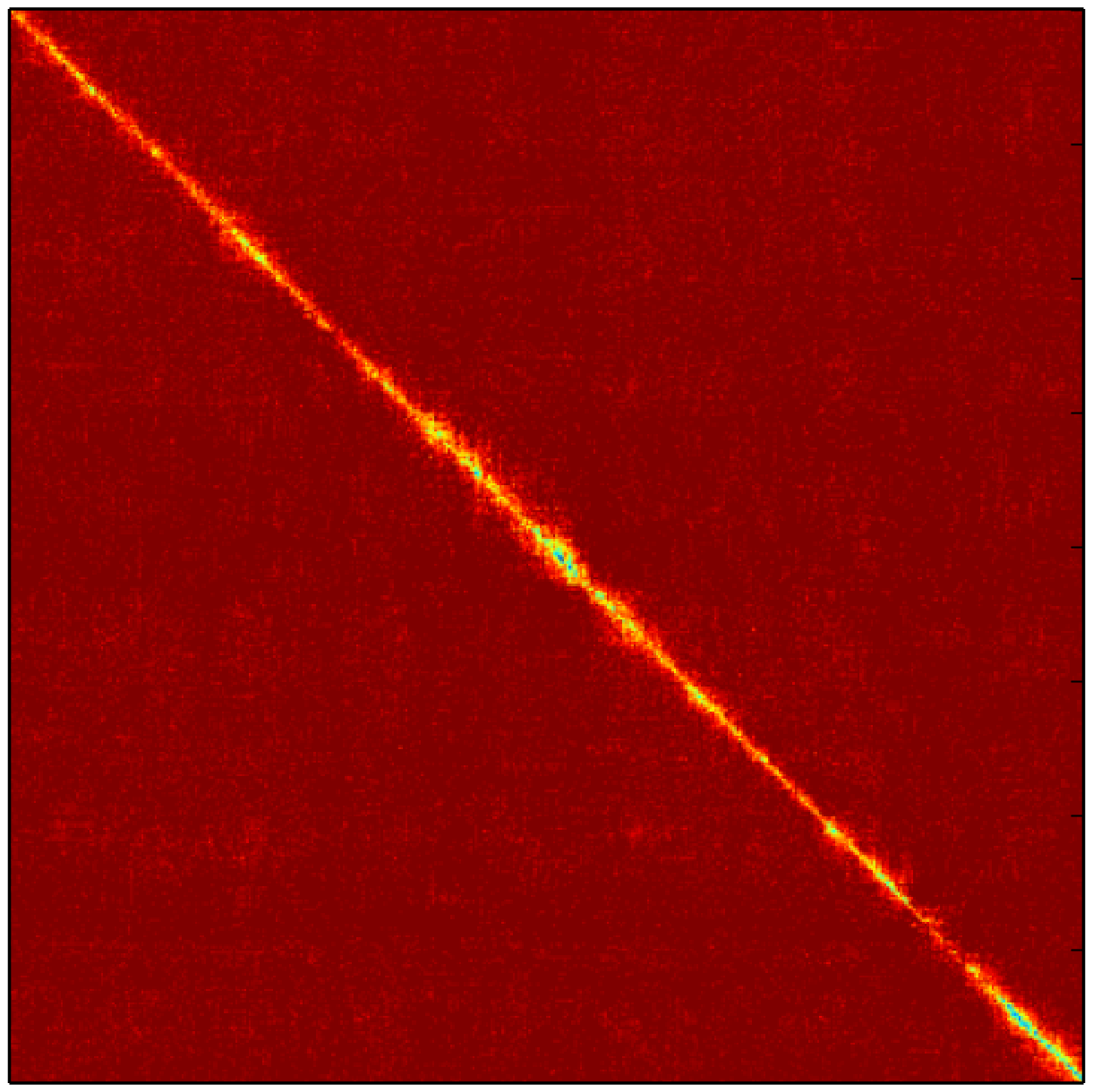}
		\vspace{\indfig}
		\label{fig_kitticonfmat_b}
		}	
	~
	\subfigure[\gcnn]{	
		\includegraphics[width=\wkitticonfmat]{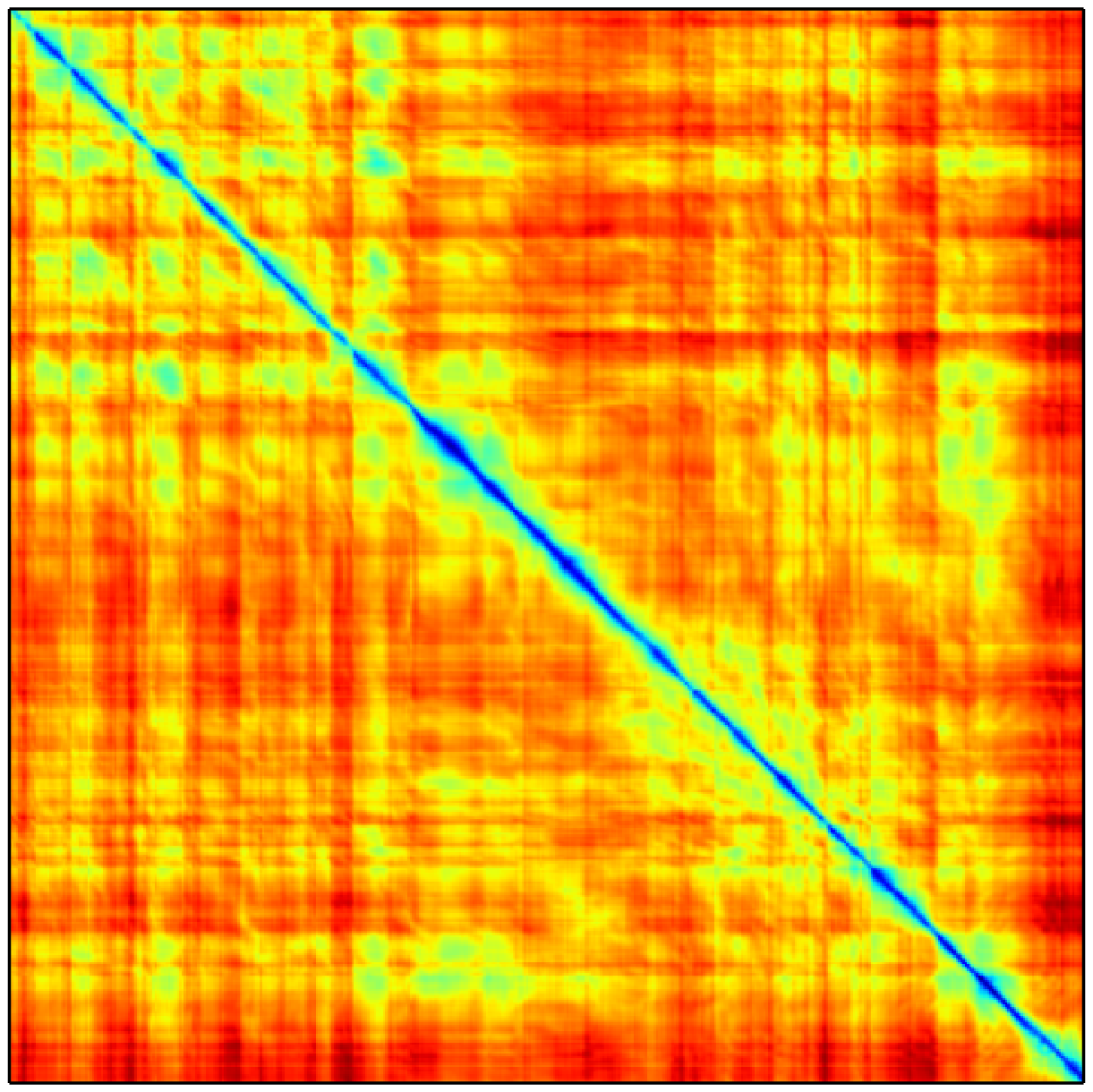}
		\vspace{\indfig}
		\label{fig_kitticonfmat_c}
		}
  	\caption{ 
	Confusion matrices belonging to our approach \ref{fig_kitticonfmat_a}, DBoW2 \ref{fig_kitticonfmat_b}, and \gcnn \ref{fig_kitticonfmat_c} in the KITTI-11 sequence, with the left camera frames as database and the right one as query. Red tones indicate high values (dissimilar images), and blue indicates low values (similar pairs). We observe a good performance on both CNN-based methods, with the difference that our method produces low distances for images with considerable changes in viewpoint, thus generating more correspondences with medium-valued Euclidean distances. In contrast, DBoW2 presents a rigid behavior, since it is unable to detect matches with large variations in viewpoint.
	} 
	\vspace{\indfig}
  	\label{fig_kitticonfmat}
\end{figure*}

\begin{figure*}[!htb]
	\centering
	\subfigure[Our approach]{	
		\includegraphics[width=\wkitticonfbin]{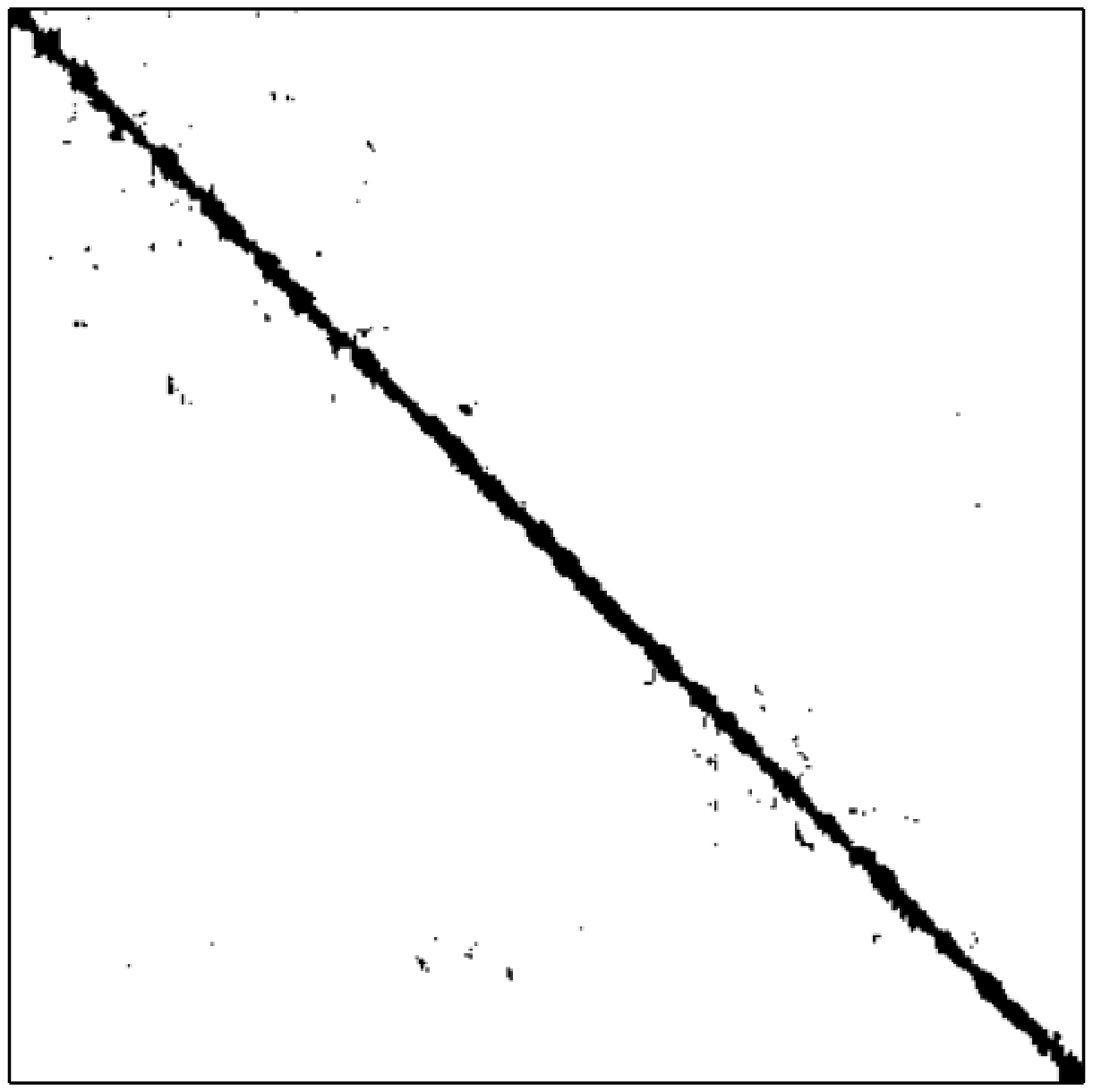}
		\vspace{\indfig}
		\label{fig_kittiknn_a}
		}
	~
	\subfigure[DBoW2]{	
		\includegraphics[width=\wkitticonfbin]{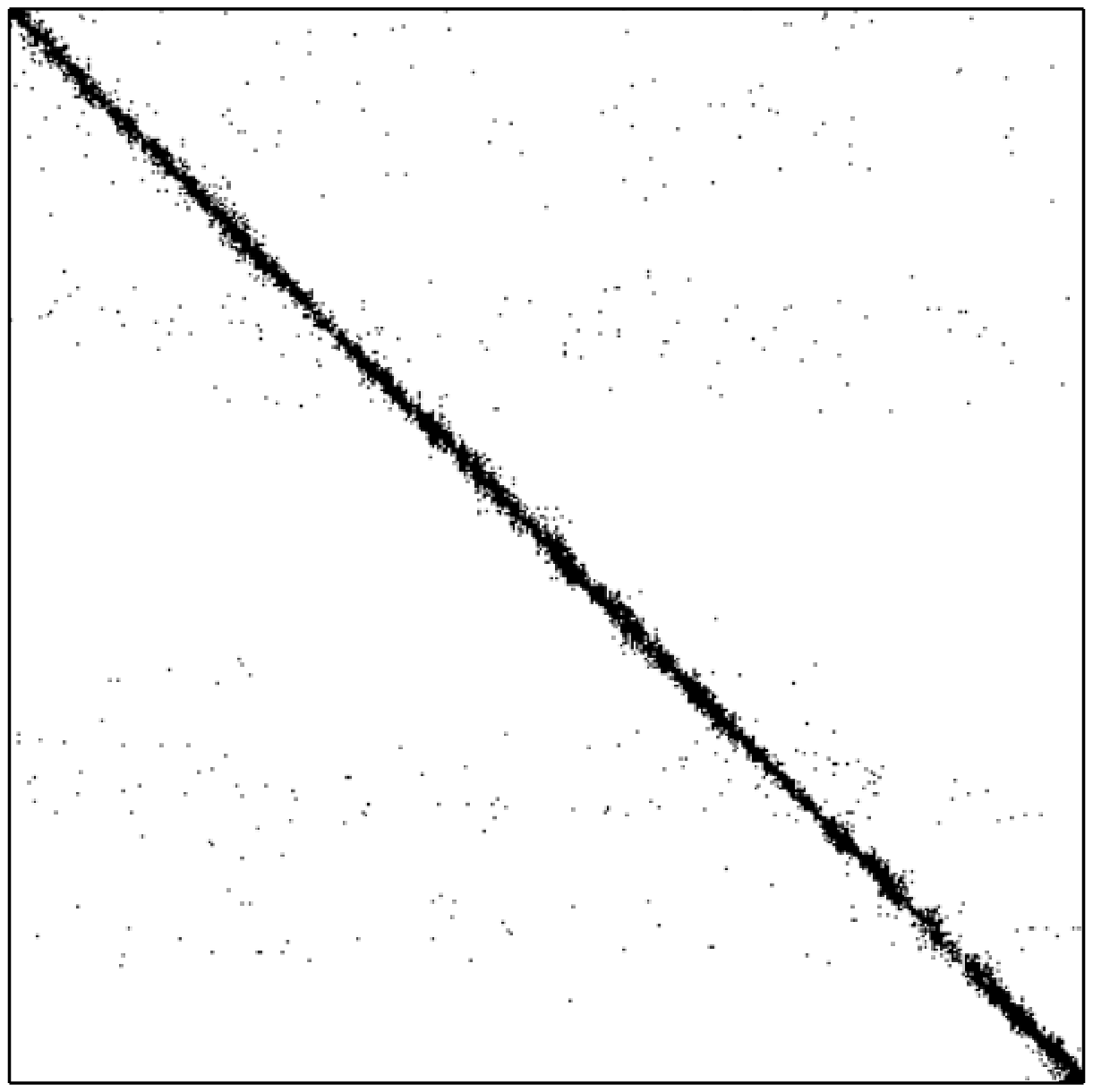}
		\vspace{\indfig}
		\label{fig_kittiknn_b}
		}	
	~
	\subfigure[\gcnn]{	
		\includegraphics[width=\wkitticonfbin]{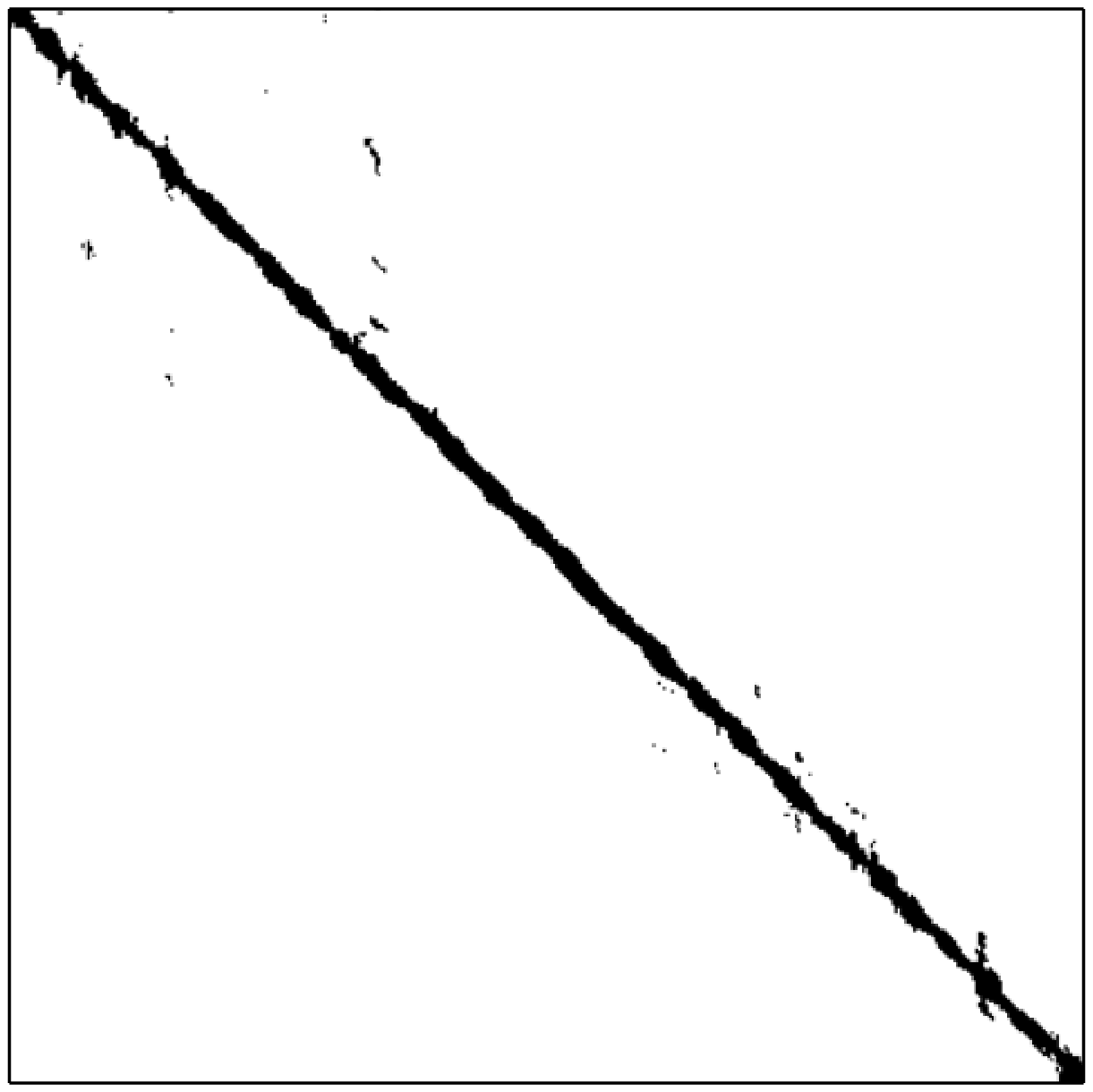}
		\vspace{\indfig}
		\label{fig_kittiknn_c}
		}
  	\caption{ 
	Binarized confusion matrices comparing the left and right images from the KITTI-11 sequence, including the $k=10$ minimum values for each query image corresponding to our approach \ref{fig_kitticonfmat_a}, DBoW2 \ref{fig_kitticonfmat_b}, and \gcnn \ref{fig_kitticonfmat_c}. The performance of the three methods is good in general, with a well-defined diagonal, but DBoW2 has a larger amount of outliers.
	} 
	\vspace{\indfig}
  	\label{fig_kittiknn}
\end{figure*}

\begin{figure}[!htb]
	\centering
	\subfigure{	
		\includegraphics[width=\wkitticurv]{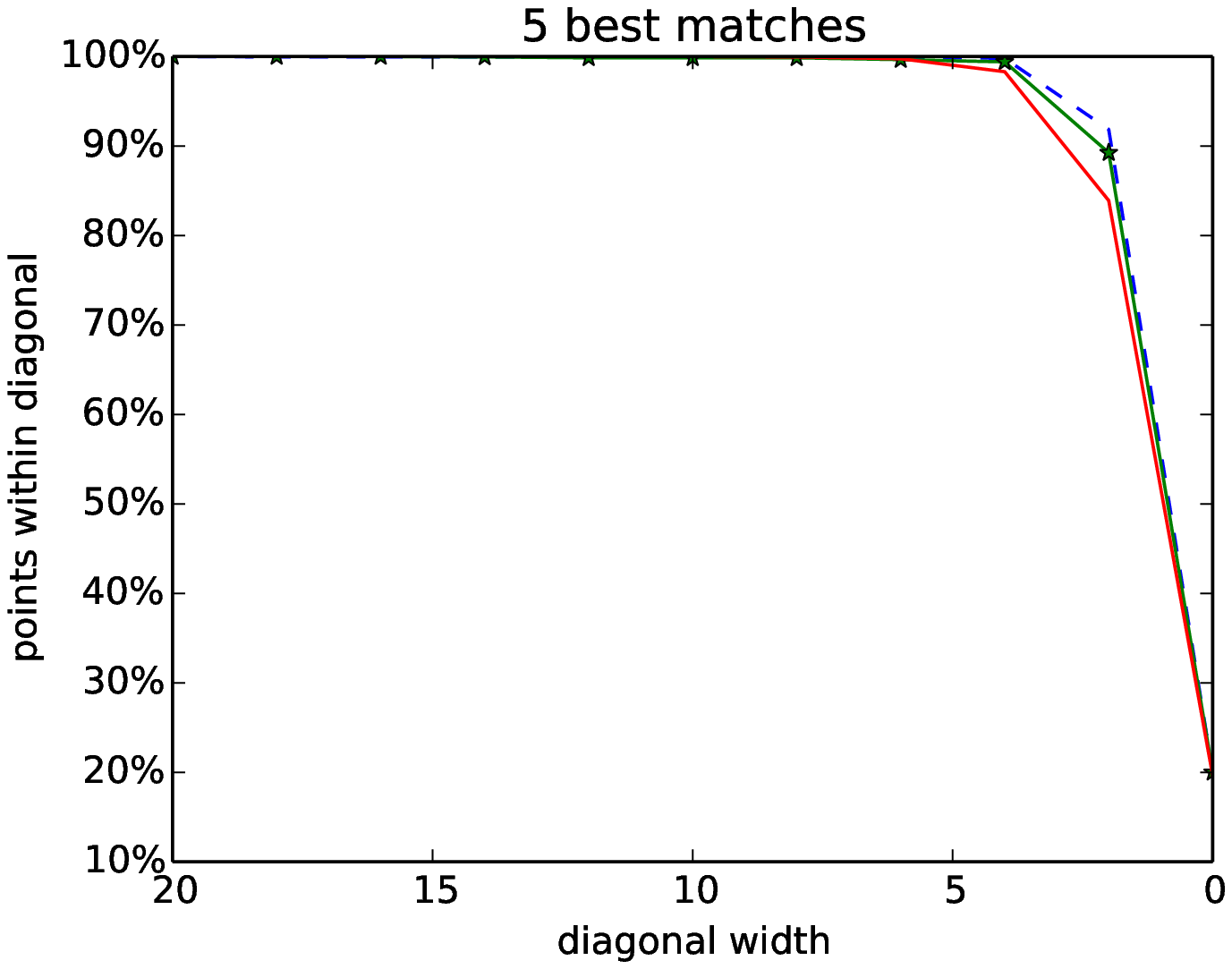}
		\vspace{\indfig}
		\label{fig_kitticurv_a}
		}
	~
	\subfigure{	
		\includegraphics[width=\wkitticurv]{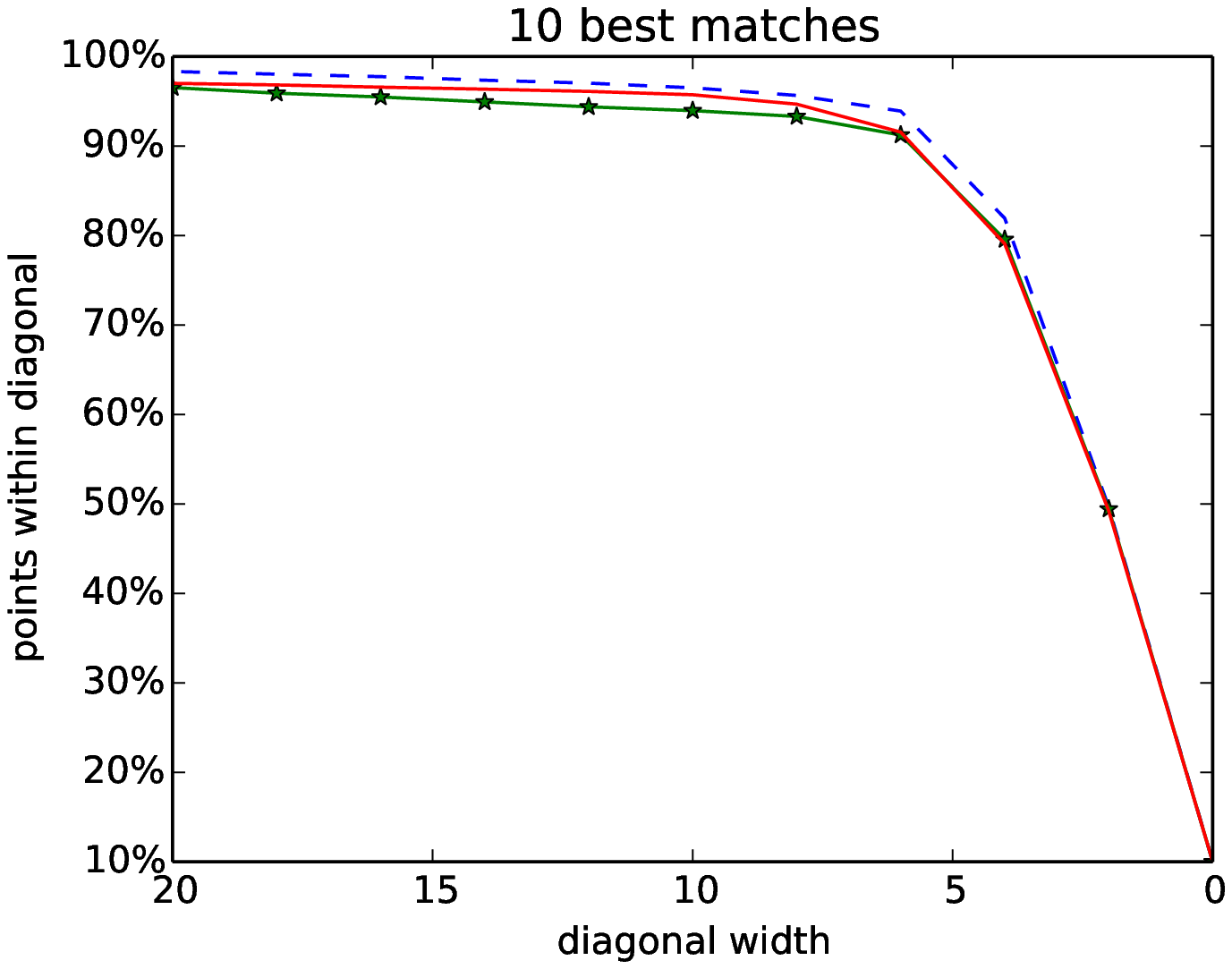}
		\label{fig_kitticurv_b}
		}	
	~
	\subfigure{	
		\hspace{\indleg}
		\includegraphics[width=\wlegend]{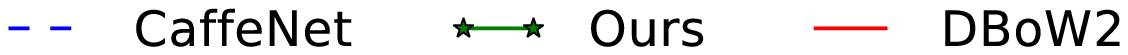}
		}	
  	\caption{ 
	Performance curves of our approach, DBoW2 and \gcnn in the KITTI-11 sequence, when comparing the left and the right sequences.
	} 
	\vspace{\indfig}
  	\label{fig_kitticurv}
\end{figure}

First, we compare the performance of the state-of-art algorithms with our proposal by processing the test sequences from the KITTI dataset \cite{geiger2012we}, which has a resolution of $1241 \times 376$.  \fig{fig_kitticonfmat} depicts the confusion matrices obtained with the sequence KITTI-11 by comparing images from the left and right cameras, where we can observe a good performance of all methods. 
We also notice that both DBoW2 and \gcnn present a thin diagonal, and they do not show any good matches outside the diagonal. 
In contrast, the confusion matrix obtained with our approach presents a thicker diagonal, and also zones with low values which correspond to parts of the sequence where the car is either stopped or circulating with low speed. 
It implies that our approach is more robust to changes in point of view, and hence, is a more versatile option for place recognition tasks which may not require the camera to be in the exact same location. 
Nevertheless, it is quite difficult to extract quantitative conclusions with the observation of these charts. 
Instead, \fig{fig_kittiknn} depicts the 10 best matches for each input image. While both \gcnn and our approach exhibit a good performance, with low dispersion around the diagonal, DBoW2 presents a considerable amount of outliers during the whole sequence. 
This is quantified in \fig{fig_kitticurv}, where we can observe that both 
CNN-based methods yield better results than DBoW2, while we observe a 
slightly superior performance of \gcnn against our approach. However,
it is worth considering that the features extracted from \gcnn are 64k-dimensional, whereas ours are much smaller, of 128 elements. This is of importance for sustainable long running place recognition as will be discussed in \secref{labtiman}.

\subsection{M\'alaga Urban Dataset}
\begin{figure}[!ht]
  \centering
    \includegraphics[width=\wmalaga]{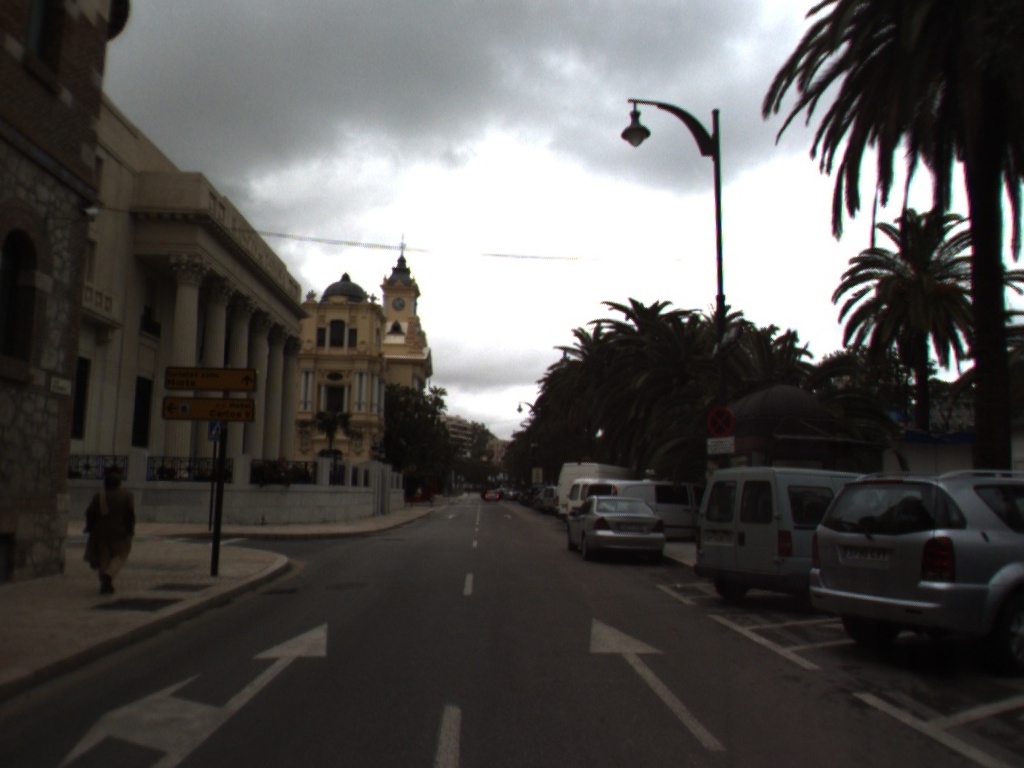}
    \caption{
	Frame extracted from the M\'alaga Urban Dataset \cite{Blanco-Claraco2014}.
    }
    \vspace{\indfig}
    \label{fig_malaga}
\end{figure}

\begin{figure}[!htb]
	\centering
	\subfigure{	
		\includegraphics[width=\wmalagacurv]{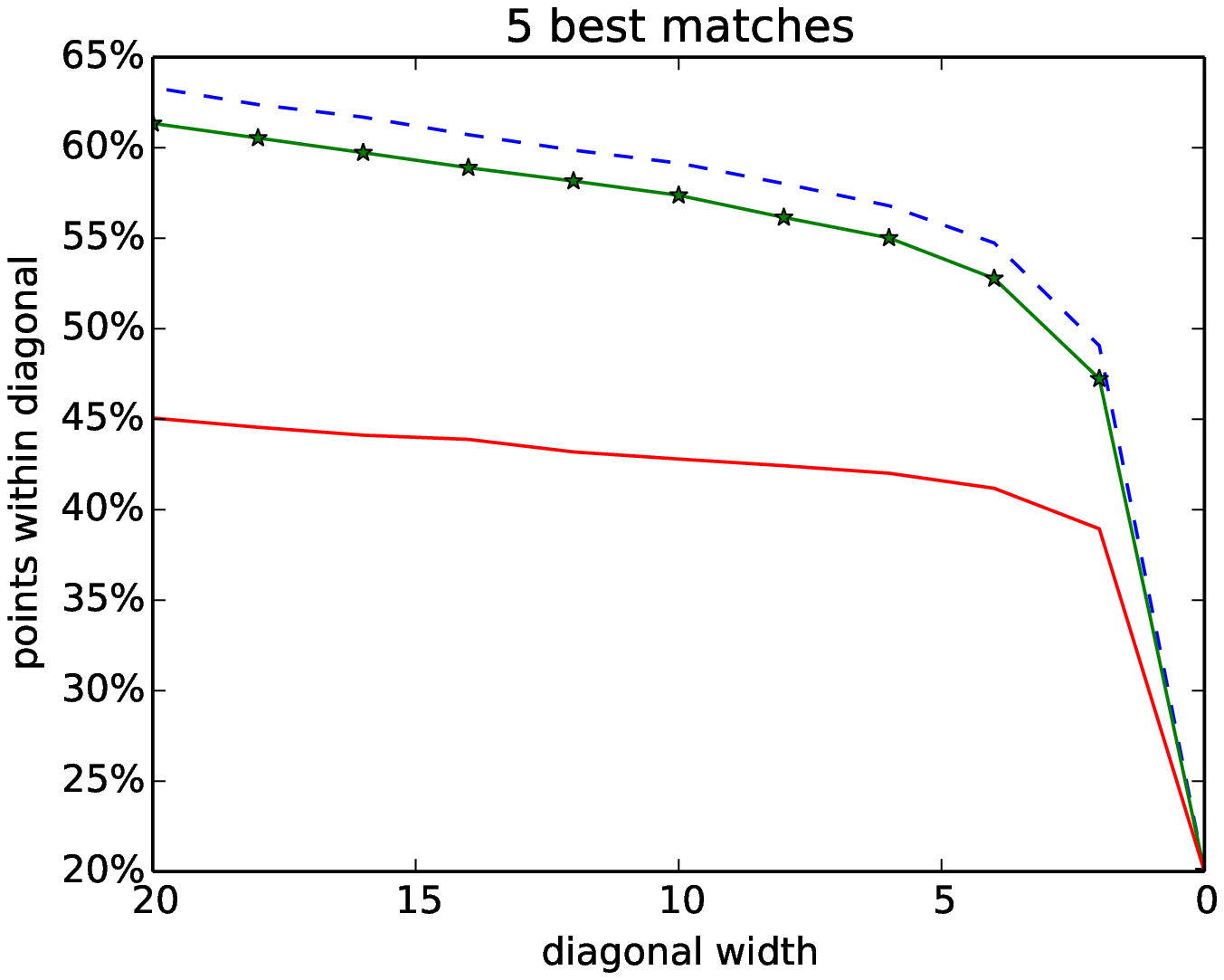}
		\vspace{\indfig}
		\label{fig_malagacurv_a}
		}
	~
	\subfigure{	
		\includegraphics[width=\wmalagacurv]{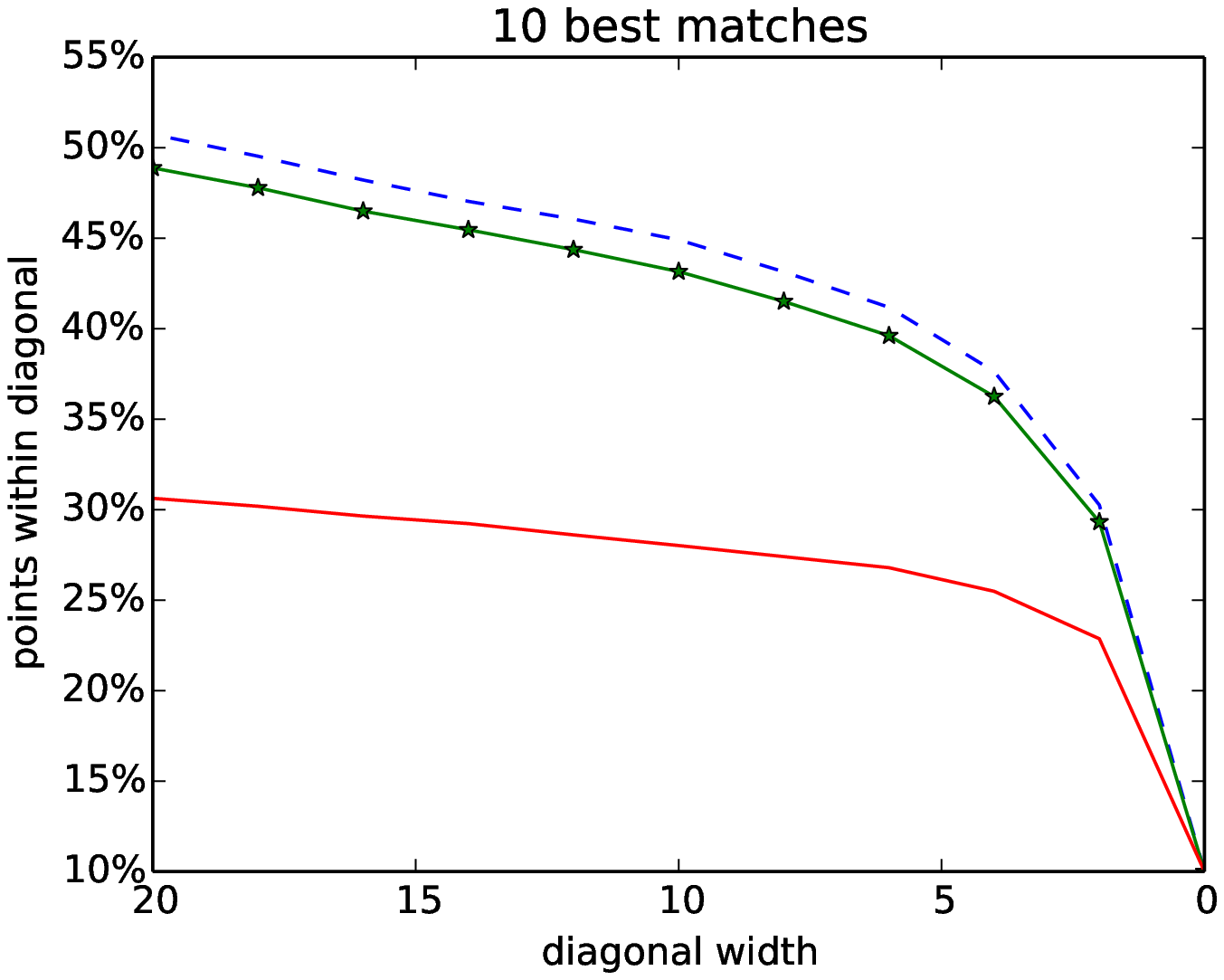}
		\vspace{\indfig}
		\label{fig_malagacurv_b}
		}
		~
	\subfigure{	
		\hspace{\indleg}
		\includegraphics[width=\wlegend]{./figures/legend_.eps}
		}		
  	\caption{ 
	Performance curves of the three methods in the M\'alaga-10 sequence, with the left and the right images as database and query inputs, respectively. 	
	} 
	\vspace{\indfig}
  	\label{fig_malagacurv}
\end{figure}

We also evaluate the performance of the techniques with the M\'alaga Urban Dataset \cite{Blanco-Claraco2014}, which contains frames obtained from a stereo camera, with a resolution of $1024 \times 768$, and data acquired from five laser scanners during a 37 km sequence in M\'alaga (Spain) with cloudy weather and direct sunlight in several parts of the sequence. 
As can be observed in \fig{fig_malaga}, the urban structure presented by this dataset is quite different than the one in the KITTI sequences, which makes it a challenging environment since none of the methods have been trained with this dataset.
\fig{fig_malagacurv} depicts the performance of the three compared methods when tested on the M\'alaga-10 (we employ the left sequence as database and the right one as query). While the CNN-based methods perform well, with a small superiority of CaffeNet, DBoW2 has a poor behavior, with a high ratio of outliers.
This proves that both CNN-based approaches are capable of recognizing images from multiple environments (even when they have not been trained with similar images), which makes them an interesting choice for life-long applications, while DBoW2 approach lacks this capability. 

\subsection{Nordland Dataset}
\begin{figure}[!htb]
	\centering
	\subfigure{	
		\includegraphics[width=\wnordcurv]{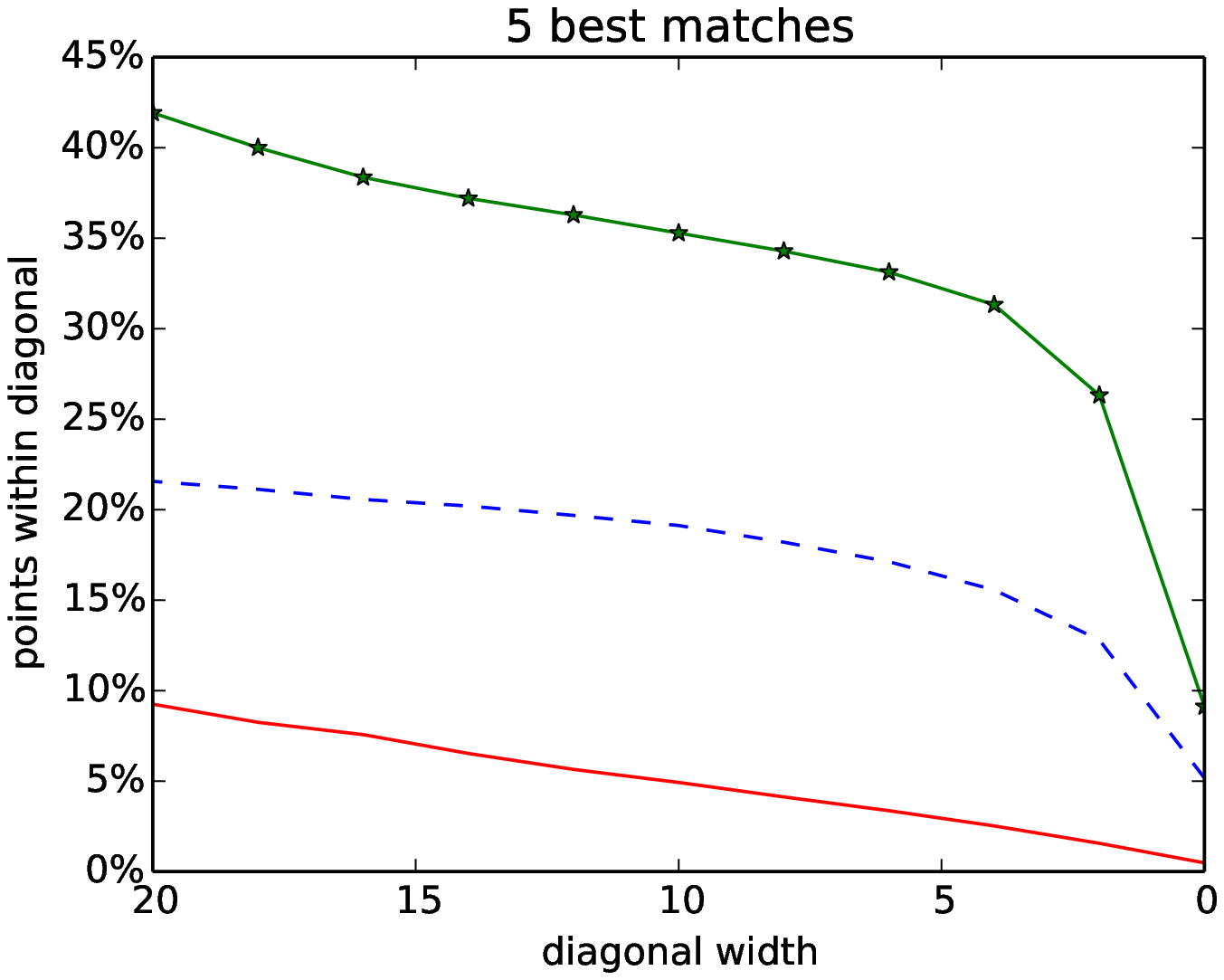}
		\vspace{\indfig}
		\label{fig_nordcurv_a}
		}
	~
	\subfigure{	
		\includegraphics[width=\wnordcurv]{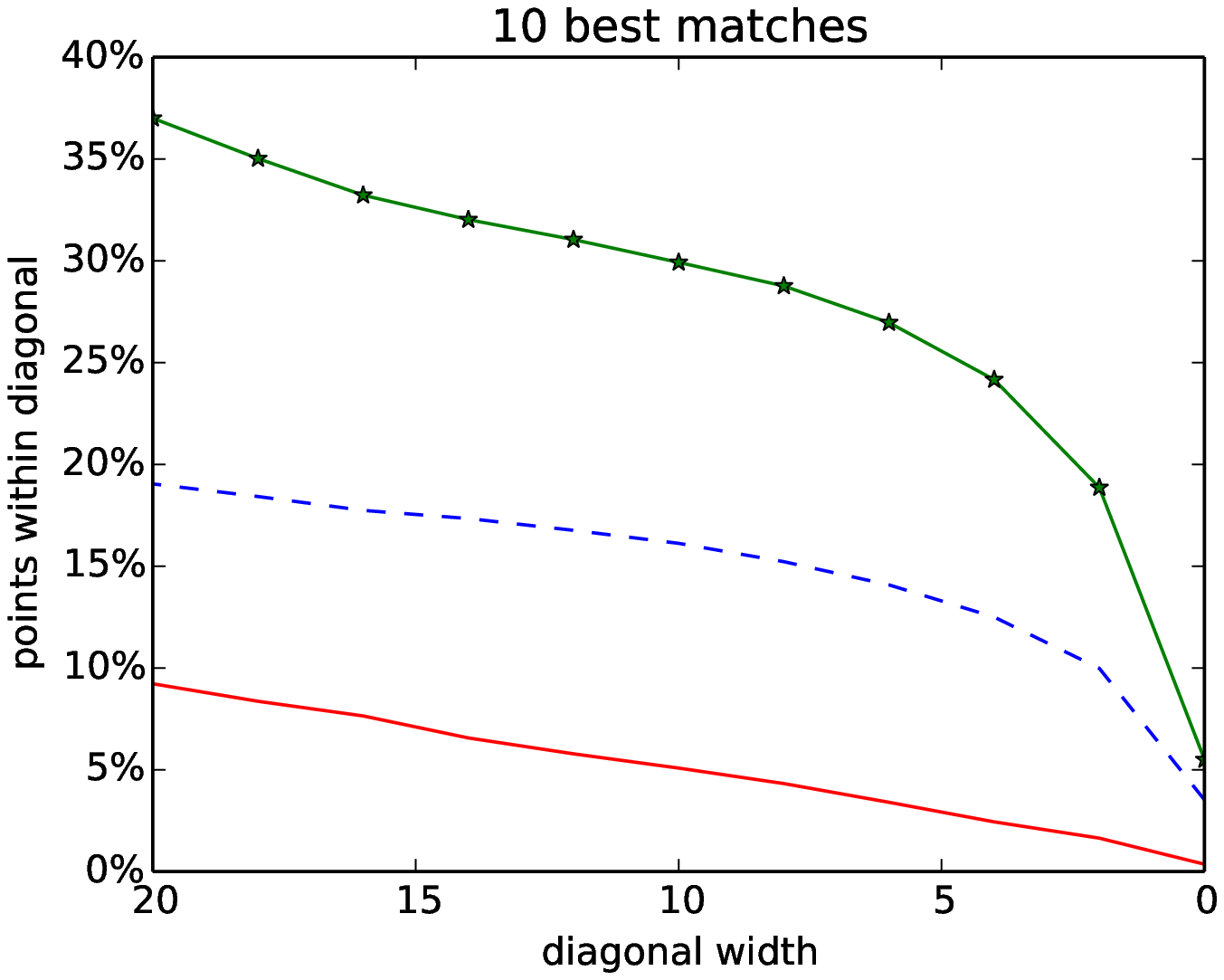}
		\vspace{\indfig}
		\label{fig_nordcurv_b}
		}
	~
	\subfigure{	
		\hspace{\indleg}
		\includegraphics[width=\wlegend]{./figures/legend_.eps}
		}		
  	\caption{ 
	Performance curves on a subset of the Nordland dataset, when using the summer sequence as database and the winter sequence as query inputs. 	
	} 
	\vspace{\indfig}
  	\label{fig_nordcurv}
\end{figure}

As mentioned above, the Nordland Dataset \cite{Sunderhauf2013} includes sequences with $1920 \times 1080$ resolution from the same perspective during the four seasons of the year, which leads to severe changes in the appearance of the environment. 
For these experiments, we have employed the last hour of the dataset, which was not used for training, and removed the segments which include either tunnels or stations.
\fig{fig_nordcurv} depicts the performance curves of the three approaches by comparing the most challenging sequence pair, summer and winter (other seasonal combinations yield similar results).
We observe the better performance of our proposal against \gcnn, which presents considerably less inliers than our approach for all the diagonal widths, for both $k=5$ and $k=10$, which is logical since neither \gcnn or DBoW2 have been trained with the purpose of being robust to those appearance changes.

\subsection{Alderley Dataset}
\begin{figure}[!htb]
	\centering
	\subfigure{	
		\includegraphics[width=\waldcurv]{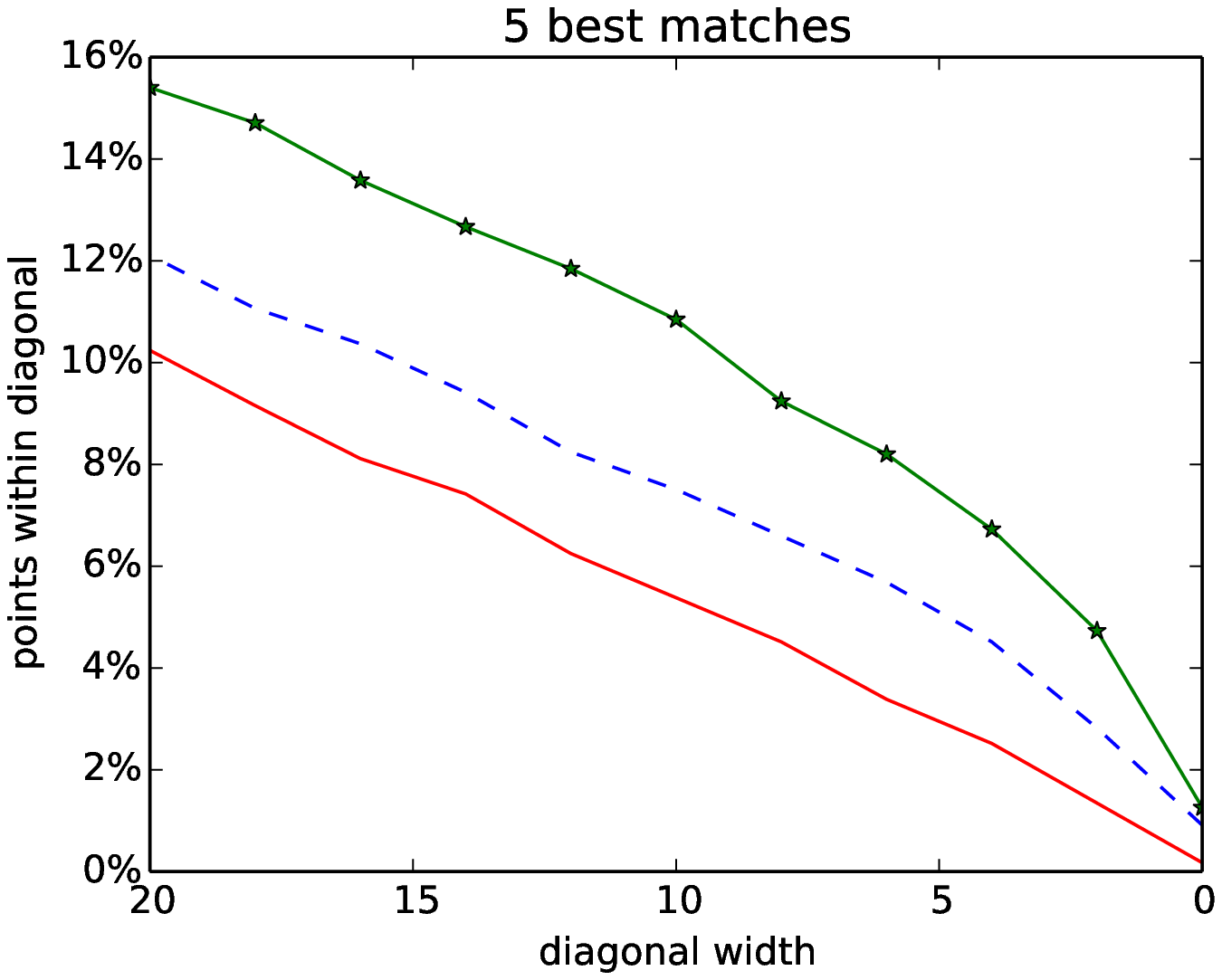}
		\vspace{\indfig}
		\label{fig_aldcurv_a}
		}
	~
	\subfigure{	
		\includegraphics[width=\waldcurv]{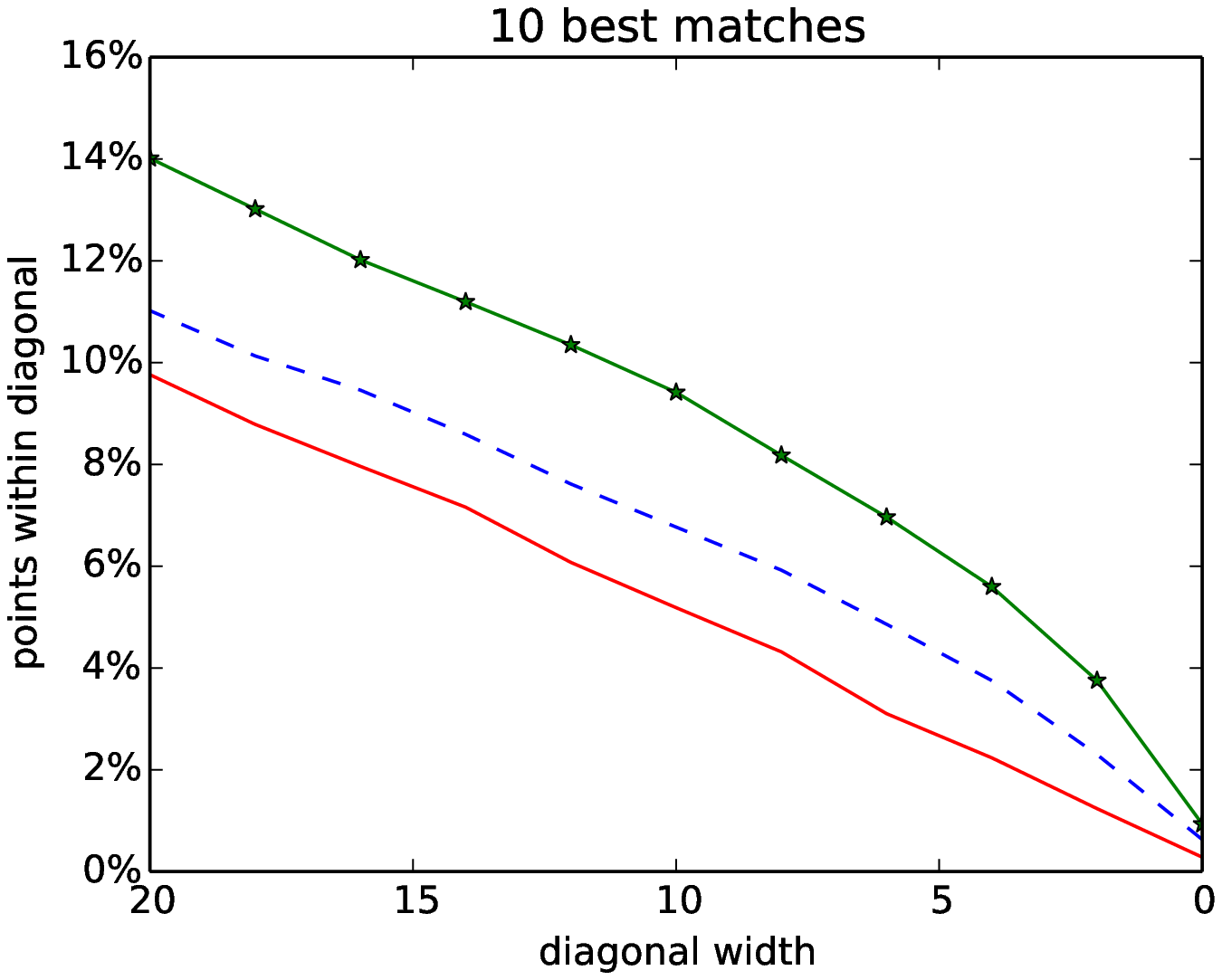}
		\vspace{\indfig}
		\label{fig_aldcurv_b}
		}
	~
	\subfigure{	
		\hspace{\indleg}
		\includegraphics[width=\waldcurv]{./figures/legend_.eps}
		}		
  	\caption{ 
	Performance curves of the three methods in the final part of the Alderley dataset. We have employed the day sequence as database, and the challenging night sequence as query.
	} 
	\vspace{\indfig}
  	\label{fig_aldcurv}
\end{figure}

We also have tested the robustness to challenging changes in weather and lightning conditions of the three methods, by processing the last 5k frames from the day sequence and their matches from the night sequence of the Alderley dataset (which has a resolution of $640 \times 260$). \fig{fig_aldcurv} shows the outperformance of our proposal against \gcnn and DBoW2, with a better ratio of inliers against diagonal width in all cases. However, it can be noticed that a low ratio is obtained by all three approaches, since it is a highly challenging dataset. Hence, the use of a post-processing technique based on sequentiality would be unavoidable to obtain a system with a reasonable performance in similar scenarios.
\subsection{Performance}
\label{labtiman}
\begin{table}
\centering
\caption {
Performance comparison between DBoW2, CaffeNet, and our approach. We compare the average processing times (in both CPU and GPU), and the descriptor lengths.
} 
\vspace{\indtab}
\label{perftable}
\begin{tabular}{@{}llll@{}} \toprule
Value                   & DBoW2    & \gcnn & Ours \\ \midrule
CPU Time (ms)          & 4-22     & 1450  & 550  \\
GPU Time (ms)          & n/a      & 30    & 10   \\
Descriptor length       & 200-500    & 64k   & 128  \\
 \toprule
\end{tabular}
\end{table}
Finally, we examine the computational performance in several 
aspects, which are presented in \tab{perftable}. 
Our tests run on an Intel Core i7-3770, while our GPU tests also rely on an NVidia 
GeForce GTX 790.
First, we measure the time required to process a single image.
In both CNN-based methods, the value
includes loading the image and performing
a forward
pass to obtain the feature vector.
In the case of DBoW2 \cite{Mur-Artal2014}, we measure the time 
required to compute the bag-of-words histogram. 
Since the input image resolution for DBoW2 is variable depending 
on the dataset, we have included the minimum and maximum average times from all the sequences.
The results indicate that DBoW2 is
less demanding than both
CNN-based methods
and that ours is three times faster than
using the reference CaffeNet network.
We then measure the size of the descriptor, which is relevant
since the computational cost of calculating the confusion 
matrix (which is required for any loop closure system)
increases with it.
The length of the word histogram of DBoW2
is variable in the official implementation,
and can be as long as the dictionary size (32k elements).
In our experiments, the length of the histogram varied from
200 to 500 elements on average, increasing in datasets
where it performs well.
On this matter, our method clearly outperforms both \gcnn and 
DBoW2 with a smaller, fixed length descriptor of 128 elements.

\section{Conclusions}
We have trained a convolutional neural network to perform place recognition under
heavy appearance changes due to weather, seasons and perspective. The network
embeds images in a 128-dimensional space where samples from similar locations
are separated by small Euclidean distances.
The network was trained using triplets of images from datasets where weather, 
lighting and point of view changes were present, in order to allow the network
to learn invariances to these changes.
The proposed network outperforms the state-of-art methods for place recognition in 
several challenging datasets, providing superior robustness to viewpoint and 
weather conditions changes. The small size of the resulting vector
makes our system suitable for applications where long-term operation is required.
\clearpage
{\small
\bibliographystyle{ieee}
\bibliography{./biblio/biblio_mendeley,./biblio/aux_biblio}
}
\end{document}